\documentclass{article}

     \usepackage[preprint,nonatbib]{neurips_2020}

\usepackage[utf8]{inputenc} 
\usepackage[T1]{fontenc}    
\usepackage{hyperref}       
\usepackage{url}            
\usepackage{booktabs}       
\usepackage{amsfonts}       
\usepackage{nicefrac}       
\usepackage{microtype}      
\usepackage{graphicx}
\usepackage{subfigure}
\usepackage[table,xcdraw]{xcolor}
\usepackage{amsmath}

\title{Policy Gradient RL Algorithms as Directed Acyclic Graphs}

\author{%
  Juan Jose Garau Luis \\
  Massachusetts Institute of Technology \\
  Department of Aeronautics and Astronautics \\
  \texttt{garau@mit.edu} \\
}

\begin{document}

\maketitle

\begin{abstract}
  Meta Reinforcement Learning (RL) methods focus on automating the design of RL algorithms that generalize to a wide range of environments. The framework introduced in \cite{coreyes2021evolving} addresses the problem by representing different RL algorithms as Directed Acyclic Graphs (DAGs), and using an evolutionary meta learner to modify these graphs and find good agent update rules. While the search language used to generate graphs in the paper serves to represent numerous already-existing RL algorithms (e.g., DQN, DDQN), it has limitations when it comes to representing Policy Gradient algorithms. In this work we try to close this gap by extending the original search language and proposing graphs for five different Policy Gradient algorithms: VPG, PPO, DDPG, TD3, and SAC.
\end{abstract}

\section{Introduction}

Novel Reinforcement Learning (RL) algorithms are recurrently being proposed to address a wide range of performance gaps. Each of these algorithms involves a human-driven design process that does not scale well. In order to reduce the design cost, the research community is focusing on meta RL algorithms, which automate the process of developing new policy update rules that generalize to multiple unseen environments. 

Broadly, meta RL methods consider a double-loop algorithm in which an agent attempts to learn good policies for arbitrary environments in an inner loop and a meta learner attempts to produce good update rules in an outer loop. The inner agent uses a loss function provided by the meta learner, whereas the meta learner uses the agent's return to guide its search.

Among all the meta RL methods that are currently being studied, in this work we focus on the framework presented in \cite{coreyes2021evolving}, in which authors introduce a search language with 26 operators used by the meta learner to form different update rules by connecting multiple operators and constructing a computation graph in the form of a Directed Acyclic Graph (DAG). The meta learner explores the space of graphs by means of evolutionary strategies and keeps track of a population with the best loss functions. The goal of the meta learner is to find the best RL algorithm given by loss function $L^*$, defined as
\begin{equation}
    L^* = \underset{L}{\text{arg max}}\left[\sum_{\varepsilon}\text{Eval}(L,\varepsilon)\right]
\end{equation}
where each $\varepsilon$ belongs to a set of different training environments, all of them used to evaluate the agent's performance, defined in turn as
\begin{equation}
    \label{eq:eval}
    \text{Eval}(L, \varepsilon) = \frac{1}{M_{\varepsilon}}\sum_{m=1}^{M_{\varepsilon}}\frac{R_m - R_{min}}{R_{max} - R_{min}}
\end{equation}
where $R_m$ corresponds to the return during episode $m$, $M_{\varepsilon}$ is the total number of episodes using environment $\varepsilon$, and $R_{min}$ and $R_{max}$ are the minimum and maximum possible returns for the environment, respectively. This method can start a search from scratch or use existing algorithms as warm-starts.

The authors test their approach using a set of training environments consisting of 4 classical control tasks from OpenAI Gym \cite{1606.01540} and 12 multitask gridworld styile environments from MiniGrid \cite{gym_minigrid}. They use 4 of these environments to meta train and 12 to meta test, and the DQN loss function \cite{mnih2015human} as an input algorithm. They find two improved versions of the DQN loss function; one of them, defined as DQNReg, outperforms both DQN and DDQN \cite{vanhasselt2015deep} in 15 out of the 16 environments. DQNReg is then tested on 4 Atari environments \cite{Bellemare_2013}, where it is also shown to outperform PPO \cite{schulman2017proximal}.

One of the salient conclusions of the paper is that representing RL algorithms as DAGs is a good way to obtain interpretable solutions. The authors reinforce this idea by including a comprehensive analysis of two of the novel update rules found by their framework based on their graph structure. While it is a promising research direction, the search language proposed in the original paper is not complete enough to build on other types of RL algorithms, such as Policy Gradient algorithms \cite{sutton2018reinforcement}. In this work we try to close this research gap by proposing extensions to their framework and representing five different Policy Gradient algorithms as DAGs: VPG, PPO, DDPG, TD3, and SAC.


\section{Related work}

Representing algorithms as computational graphs is also proposed in other studies. In \cite{real2019regularized} the same concept is used to find appropriate image classifiers. This idea is also present in \cite{alet2020metalearning}, where a meta learner produces curiosity graphs that drive the agent's exploration by shaping the environment's reward. The agent then uses the modified reward with a fixed update rule. Conversely, the evolutionary strategy in \cite{coreyes2021evolving} allows flexible update rules and is meta trained on a diverse set of environments.

As oposed to changing the form of RL algorithms, other studies consider a fixed-form update rule which takes in parameters provided by the meta learner. It is the case of \cite{houthooft2018evolved}, where authors propose an evolutionary strategy to learn good policy update parameters $\hat{\pi}$ that guide the policy optimization. They train the meta RL framework on several MuJoCo tasks \cite{Todorov2012MuJoCoAP} and test on similar versions of those tasks. The same train and test procedure is followed by the authors in \cite{bechtle2020metalearning}, although a meta-gradient descent strategy is adopted instead of evolutionary methods. This approach is also followed by the authors in \cite{kirsch2020improving}, which propose MetaGenRL to attain higher generalization capabilities. Those are proven in test environments based on unseen MuJoCo tasks.

In addition to learning appropriate policy update rules $\hat{\pi}$, the method presented in \cite{oh2020discovering}, defined as Learned Policy Gradient, also learns a prediction update rule $\hat{y}$, i.e., the semantics of the agent's prediction. In some experiments, the discovered semantics converge towards a notion of value function. The authors claim LPG is able to attain better generalizability this way, as proven on the performance on unseen Atari test environments after meta training on toy environments.

\section{Methods}

\begin{figure}[t]
\centering
\includegraphics[width=.98\linewidth]{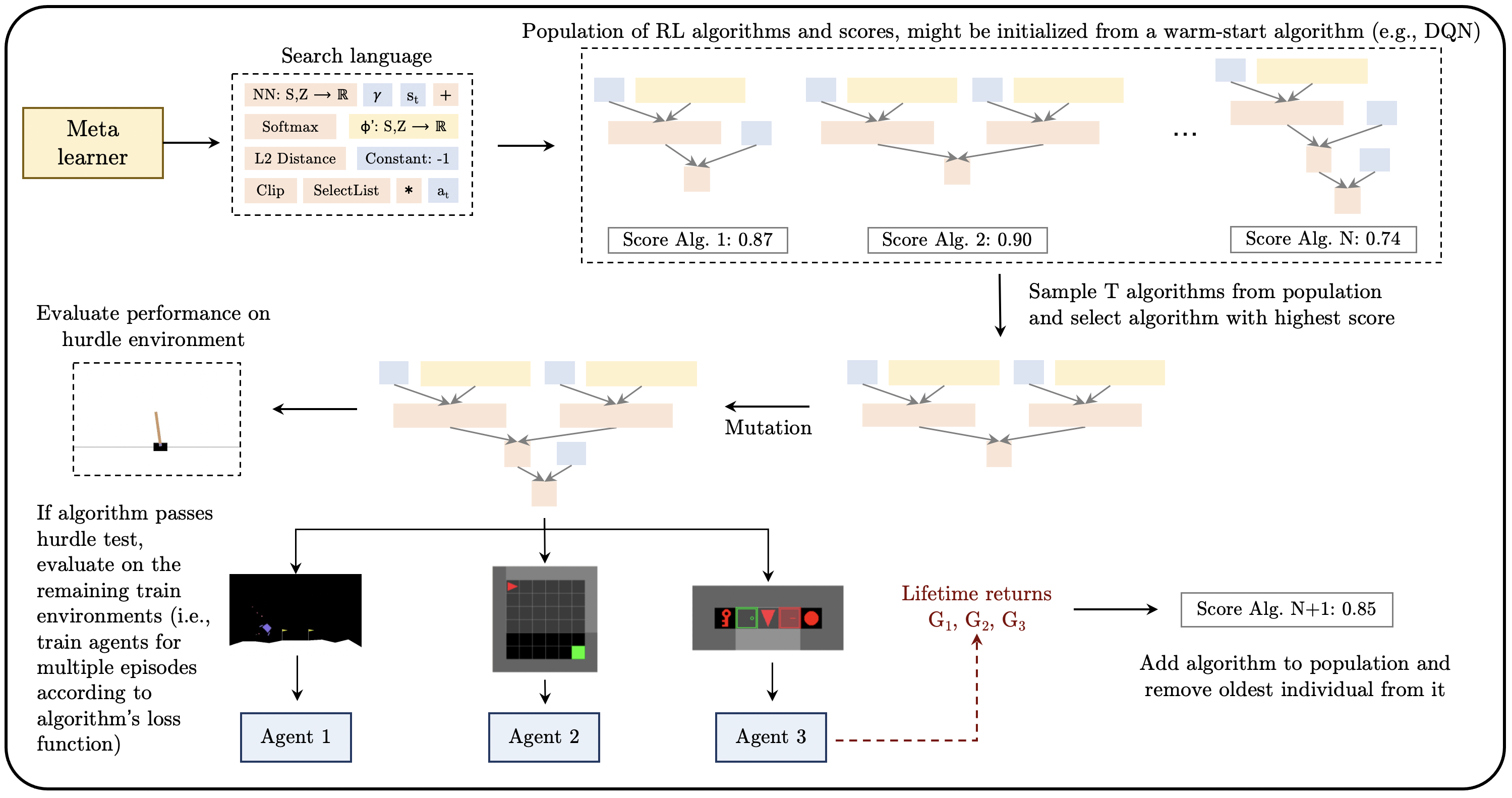}
\caption{Evolutionary Reinforcement Learning method \cite{coreyes2021evolving} overview. The meta learning framework constructs a population of RL algorithms or loss functions from a predefined search language, randomly creating each one or using an already-existing algorithm as a warm-start. For a fixed number of iterations, the best RL algorithm from a random set of $T$ algorithms undergoes mutation and a child algorithm is generated. This new algorithm is evaluated, first on a hurdle environment, and then on a set of training environments. The meta learner collects the returns from each environment and computes the score of the new algorithm. The oldest algorithm from the population is then removed.}
\label{fig:evo_arch}
\end{figure}

The focus of this work is to extend the applicability of the framework presented in \cite{coreyes2021evolving} by adapting the search language to Policy Gradient algorithms \cite{sutton2018reinforcement}. First, we present a high-level description of the method from the original paper, summarized in Figure \ref{fig:evo_arch}. Then, we introduce our extensions to the search language.

\subsection{General Description}

The input to the algorithm consists of a set of training environments, one additional ``hurdle environment'', and a search language that conforms the nodes of the computational graphs. The algorithm starts by creating a population of $N$ \footnote{This symbol is defined by us, it is not in the original paper} RL algorithms or loss functions, which can be constructed from scratch using the search language or from an input already-existing RL algorithm. The starting population is scored in each of the training environments.

For a defined number of $C$ iterations, the meta learner first samples a set of $T$ algorithms from the population and selects the one with the highest score as a parent algorithm. Then, a child algorithm is created by mutating the parent algorithm. The hurdle environment is used to assess a minimum performance threshold $\alpha$, which determines whether the child algorithm is considered for the population or not. If it passes the threshold, the child algorithm is evaluated on each of the training environments by independent agents, and an aggregated score is generated. The new algorithm is added to the population, which is kept to a constant size of $N$ individuals by removing the oldest one.

\subsection{Search Language Extensions}

\begin{figure}[!h]
\centering
\includegraphics[width=.5\linewidth]{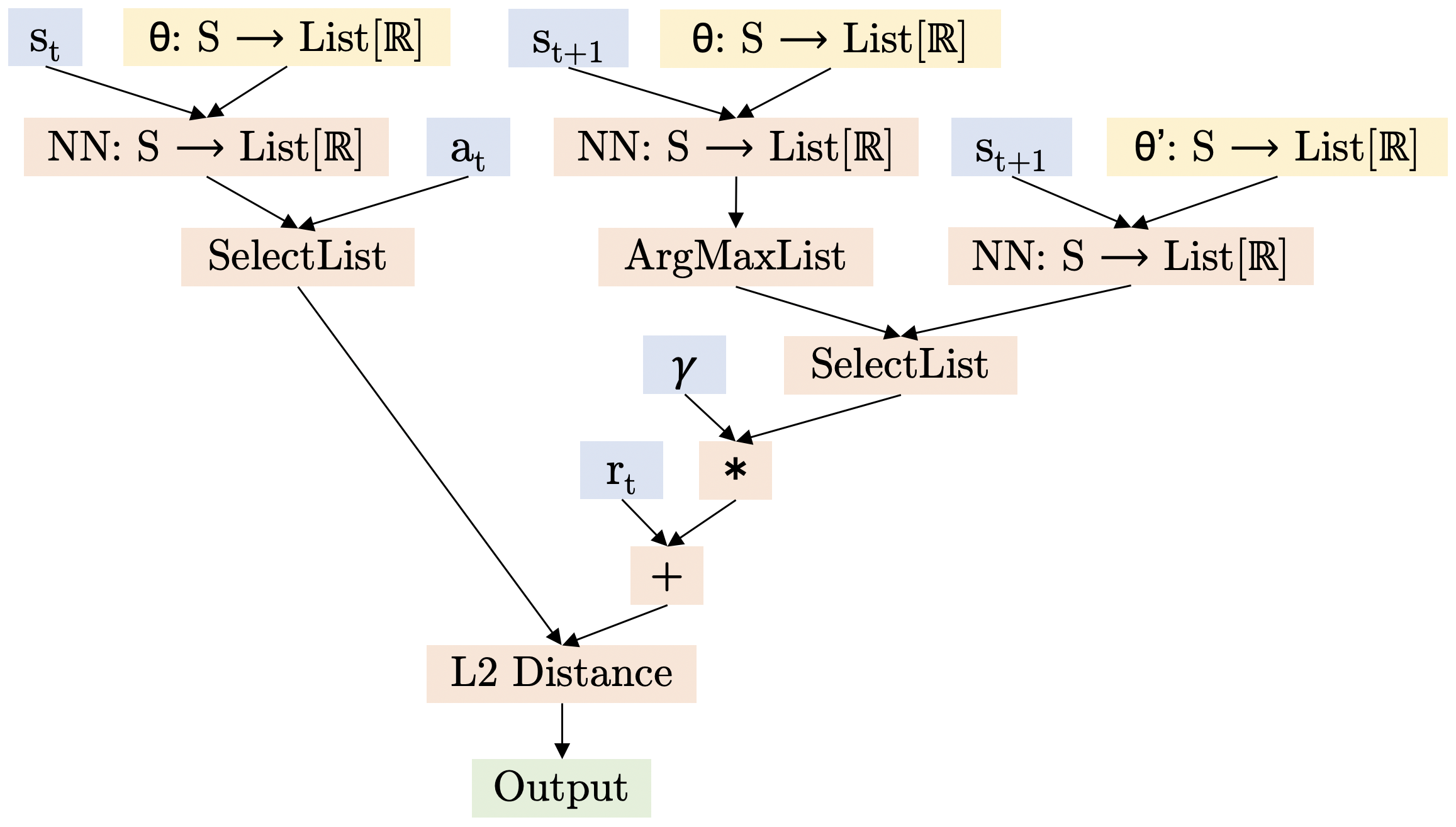}
\caption{Computational graph representing DDQN loss function $L_{\theta} = \left(Q_{\theta}\left(s_t, a_t\right) - \left(r_t + \gamma Q_{\theta'}\left(s_{t+1}, \underset{a}{\text{arg max}} Q_{\theta}\left(s_{t+1}, a\right)\right)\right)\right)^2$.}
\label{fig:ddqn}
\end{figure}

The framework expresses RL algorithms as DAGs of nodes with typed inputs and outputs. The input nodes consist of the elements in the tuple $(s_t, a_t, r_t, s_{t+1})$, the network parameters $\theta$, the target network parameters $\theta'$, the discount factor $\gamma$, and other numerical constants (e.g., 1, 0.1). Other types of nodes include parameter nodes, which encode neural network operations, and operation nodes, which perform scalar and list operation over the inputs (e.g., \textit{Add}, \textit{MaxList}). A full list of the available nodes can be found in the appendix of the original paper \cite{coreyes2021evolving}. For reference, Figure \ref{fig:ddqn} shows the DDQN algorithm \cite{vanhasselt2015deep} encoded as a computational graph using the \emph{original} set of nodes from the paper.

The output of the computational graph, given by the node \textit{Output}, is then used as the loss function $L_{\theta}$ the agent must minimize, updating its parameters by gradient descent $\theta \longleftarrow \theta - \alpha_{lr}\nabla_{\theta}L_{\theta}$. We follow this gradient descent convention throughout the rest of this work.

We now introduce each of the extensions to the original language, divided into new input nodes, new operation nodes, and new constant nodes. We follow the data type notation convention proposed by the authors, which includes state $\mathbb{S}$, action $\mathbb{Z}$, float $\mathbb{R}$, and list $List[\mathbb{X}]$, where $\mathbb{X}$ indicates it can be of $\mathbb{S}$ or $\mathbb{R}$\footnote{Notation extracted from the original paper \cite{coreyes2021evolving}}.

\subsubsection{Input nodes}

In the paper it is assumed each independent agent initializes Q-value parameters $\theta$, target network parameters $\theta'$, and an empty replay buffer $\mathcal{D}$, where the agent stores all the transition tuples encountered in order to do off-policy learning. To adapt the framework to Policy Gradient algorithms, we propose the following extensions:
\begin{itemize}
    \item We register the episode-termination signal $d_t$ in the replay buffer $\mathcal{D}$.
    \item We add the parameter $\lambda$ \cite{sutton2018reinforcement} to the language.
    \item We allow the tuple $(s_t, a_t, r_t, d_t, s_{t+1})$ to represent a sequence of consecutive timesteps, instead of a single timestep. According to the original paper, operations broadcast in the cases in which inputs have a different number of dimensions.
    \item We use additional network parameters $\phi$ (e.g., for the value network when needed) and their target counterparts $\phi'$.
    \item In addition to the network parameter nodes from the original paper, which allow mappings from states $\mathbb{S}$ to floats $\mathbb{R}$ and lists $List[\mathbb{R}]$, we also allow to map both a state $\mathbb{S}$ and an action $\mathbb{Z}$ to a float $\mathbb{R}$, i.e., $\theta:\mathbb{S},\mathbb{Z}\longrightarrow\mathbb{R}$. Likewise, we introduce the parameter node $NN:\mathbb{S},\mathbb{Z}\longrightarrow\mathbb{R}$ to carry out operations of such kind. We do this in order to address \emph{continuous} state and action spaces, as required by certain algorithms, such as DDPG \cite{lillicrap2015continuous}.
\end{itemize}

\subsubsection{Operation nodes}

Apart from the operation nodes proposed in the original paper, we introduce the additional nodes:
\begin{itemize}
    \item \textit{SumAndDiscount}: It takes in a vector $v$ of length $l_v$ and a numerical value $b$, and outputs a vector $v'$ of length $l_v$ where each element is computed as $v'_i = \sum_{k=i}^{l_v} v_k \cdot b^{k - i}$. This is necessary to compute certain elements of different Policy Gradient methods, such as computing the rewards-to-go (in that case, $v = r$ and $b = 1$).
    \item \textit{Clip}: It clips a scalar or vector component-wise given minimum and maximum float values.
    \item \textit{Squashing}: Given a state $s_t$ from data type $\mathbb{S}$ and network parameters $\theta:\mathbb{S}\longrightarrow\mathbb{R}$, it computes actions $\tilde{a}$ following a squashed Gaussian policy, i.e., $\tilde{a} = \tanh(\mu_{\theta}(s_t) + \sigma_{\theta}(s_t) \odot \xi)$, with $\xi\sim\mathcal{N}(0, I)$. This node is specifically introduced to represent the Soft Actor-Critic algorithm \cite{haarnoja2018soft}.
    \item \textit{Prob}: In cases in which the action space is continuous, and therefore the node \textit{Softmax} can not be used, this node computes the probability of taking a certain action in a given state with given network parameters $\theta$. It assumes the probabilities can be computed, such as in the case of Gaussian or categorical policies. 
\end{itemize}

\subsubsection{Constant nodes}

Finally, we encode useful constant values as new constant nodes for the language. These are general values, such as $-1$, or algorithm-specific constants, such as $1+\epsilon$ and $1-\epsilon$, for the example case of PPO. We introduce all algorithm-specific constant nodes in their related sections.

\section{Example Graphs}

Based on the language extensions introduced in the previous section, we propose computational graphs in the form of DAGs for the following Policy Gradient algorithms: Vanilla Policy Gradient (VPG), Proximal Policy Optimization (PPO) \cite{schulman2017proximal}, Deep Deterministic Policy Gradient (DDPG) \cite{lillicrap2015continuous}, Twin Delayed DDPG (TD3) \cite{fujimoto2018addressing}, and Soft Actor-Critic (SAC) \cite{haarnoja2018soft}. We base our graph constructions on the descriptions and implementations provided in \cite{SpinningUp2018}. For all algorithms, we leave the target parameters update considerations (i.e., $\theta' \longleftarrow \theta$, $\phi' \longleftarrow \phi$) for the framework user to decide.

\subsection{Vanilla Policy Gradient (VPG)}

The work in \cite{coreyes2021evolving} only considers off-policy algorithms that sample batches of tuples $(s_t, a_t, r_t, s_{t+1})$. In this work we also allow $s_t$ to represent a sequence of consecutive states. We apply this idea to construct a computational graph for the VPG algorithm that can be used as an input to the meta learning framework. We need to consider two different loss functions: one for the policy $\pi_{\theta}$ and another for the value function $V_{\phi}$. 

\begin{figure}[!h]
\centering
\includegraphics[width=.5\linewidth]{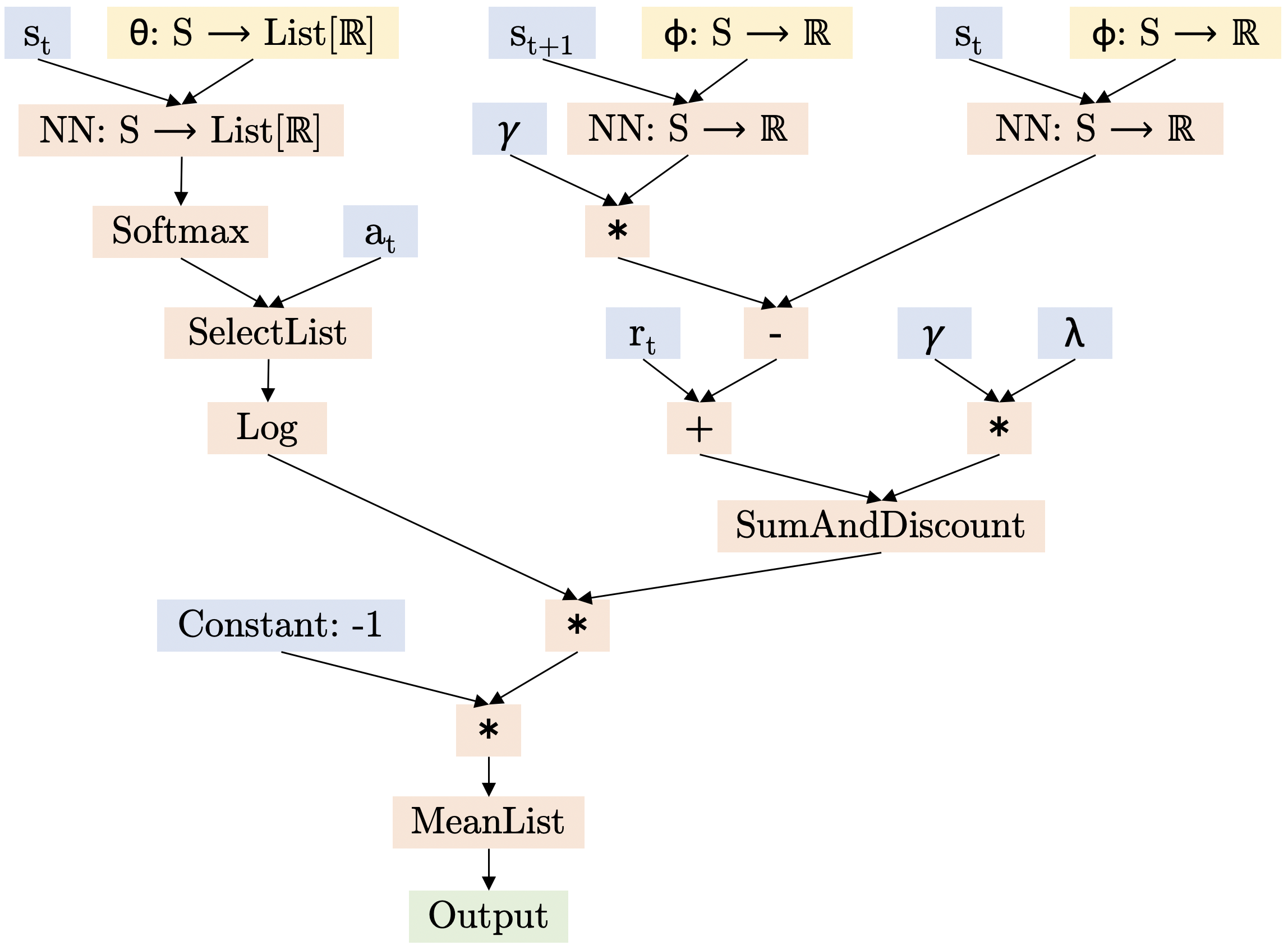}
\caption{Computational graph representing Vanilla Policy Gradient (VPG) policy $\pi_{\theta}$ loss $L_{\theta} = -\frac{1}{T}\sum_{t=0}^T\log\pi_{\theta}\left(a_t|s_t\right)\hat{A}_t$.}
\label{fig:vpg_pi}
\end{figure}

First, Figure \ref{fig:vpg_pi} shows the computational graph for the policy loss, in which we assume the use of GAE-Lambda advantage estimation method \cite{sutton2018reinforcement} to compute the discounted advantage for each timestep. In this example, we introduce the novel node \textit{SumAndDiscount} and a constant node with the value $-1$, in order to follow the loss minimization convention from the original paper. 

\begin{figure}[!h]
\centering
\includegraphics[width=.28\linewidth]{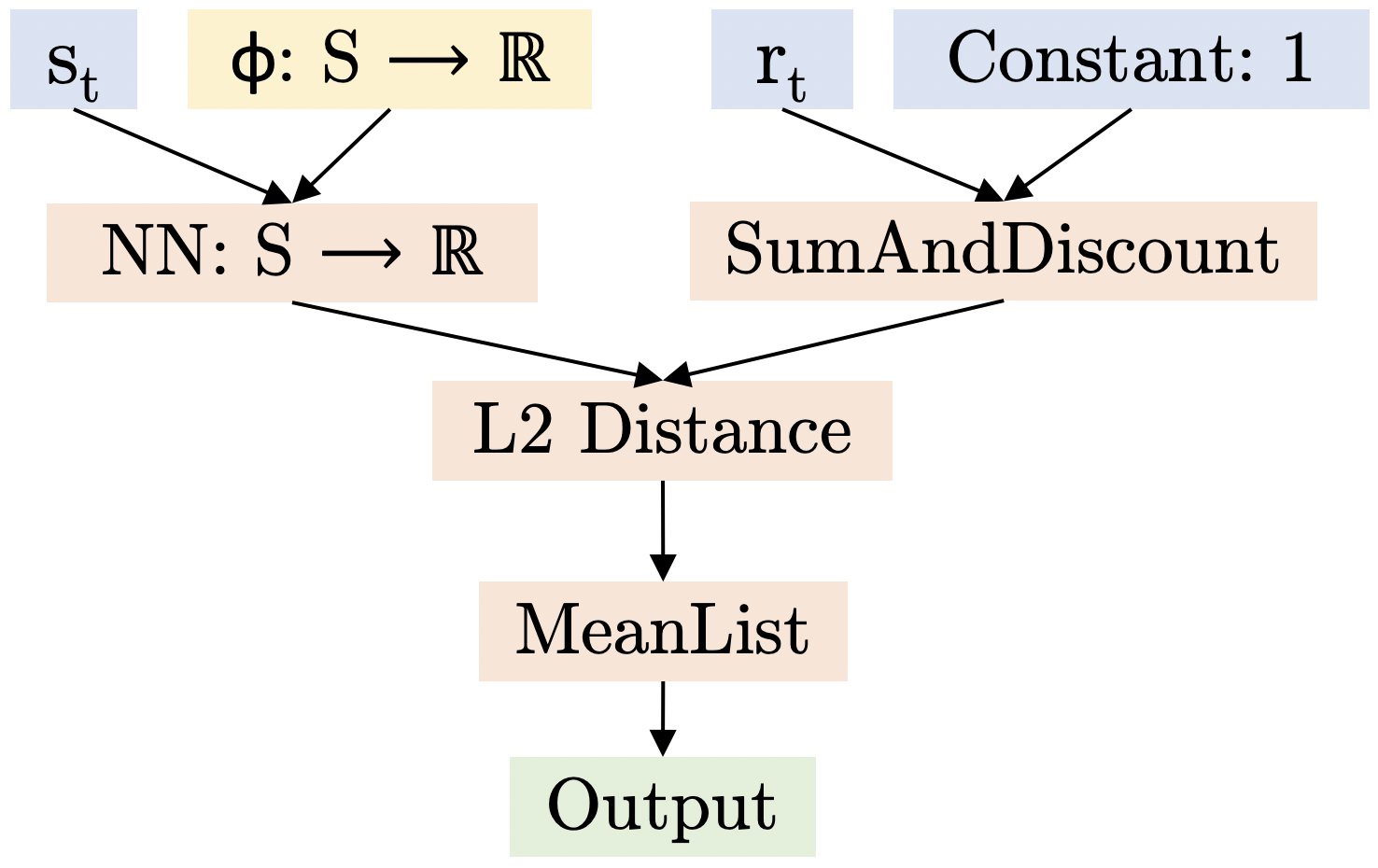}
\caption{Computational graph representing Vanilla Policy Gradient (VPG) value function $V_{\phi}$ loss $L_{\phi} = \frac{1}{T}\sum_{t=0}^T\left(V_{\phi}(s_t) - \hat{R}_t\right)^2$.}
\label{fig:vpg_v}
\end{figure}

Then, Figure \ref{fig:vpg_v} shows the computational graph for the value function loss, which computes the MSE between the value function and the rewards-to-go.

\subsection{Proximal Policy Optimization (PPO)}

\begin{figure}[!h]
\centering
\includegraphics[width=.57\linewidth]{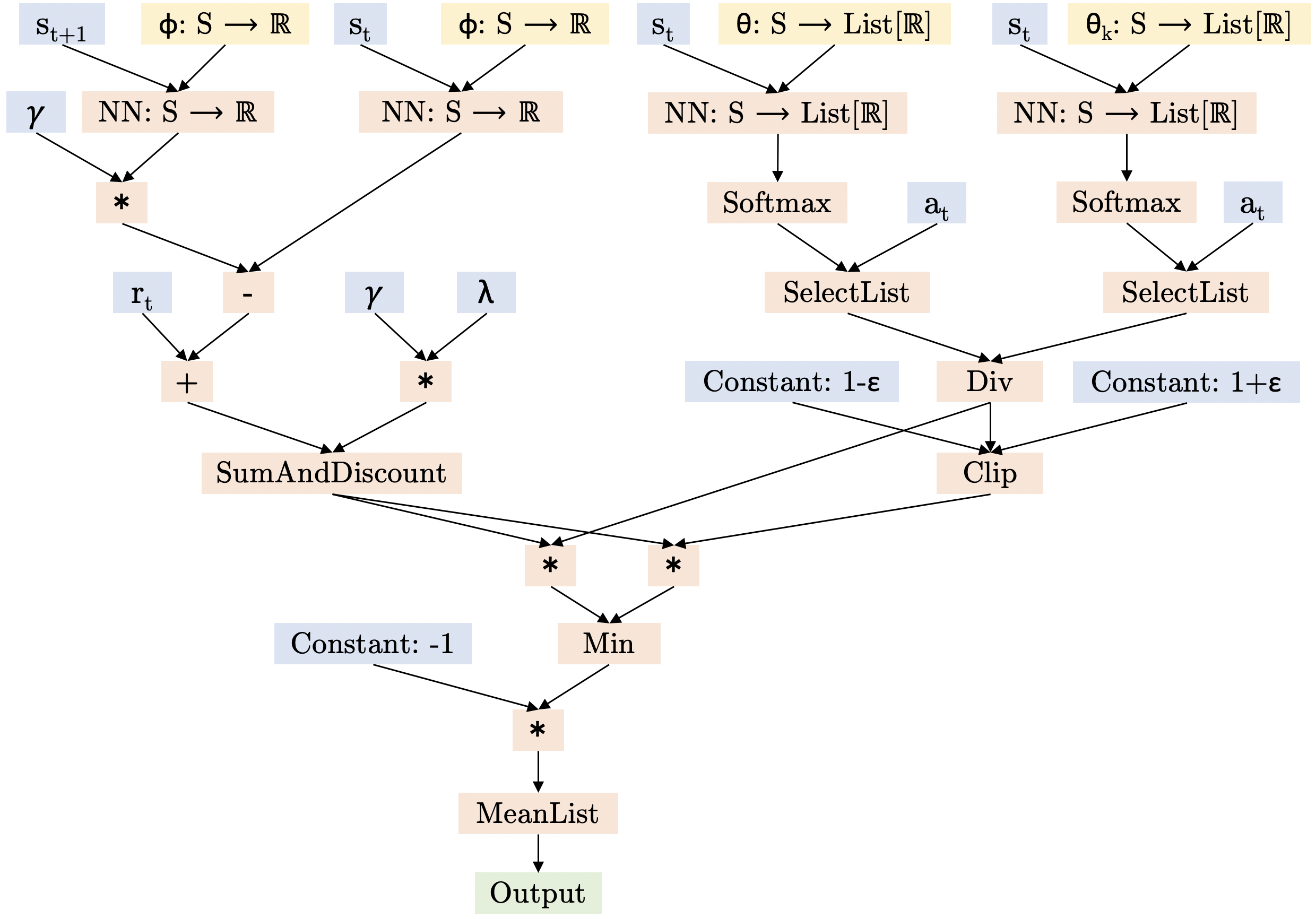}
\caption{Computational graph representing Proximal Policy Optimization (PPO) policy $\pi_{\theta}$ loss $L_{\theta} = - \frac{1}{T}\sum_{t=0}^T\min\left(\frac{\pi_{\theta}(a_t|s_t)}{\pi_{\theta_k}(a_t|s_t)}\hat{A}_t, \text{clip}\left(\frac{\pi_{\theta}(a_t|s_t)}{\pi_{\theta_k}(a_t|s_t)}, 1 - \epsilon, 1 + \epsilon\right)\hat{A}_t\right)$.}
\label{fig:ppo}
\end{figure}

In the case of the PPO algorithm we consider the policy parameters $\theta$ as well as an early version $\theta_k$ to compute the policy change ratio. Figure \ref{fig:ppo} shows the computational graph for this algorithm, which includes the novel node \textit{Clip}, as well as the constant nodes $1+\epsilon$ and $1-\epsilon$, input to the clipping function. We use the same advantage estimation method used in the VPG case, and also consider the same value function update method, already presented in Figure \ref{fig:vpg_v}.

\subsection{Deep Deterministic Policy Gradient (DDPG)}

\begin{figure}[!h]
\centering
\includegraphics[width=.55\linewidth]{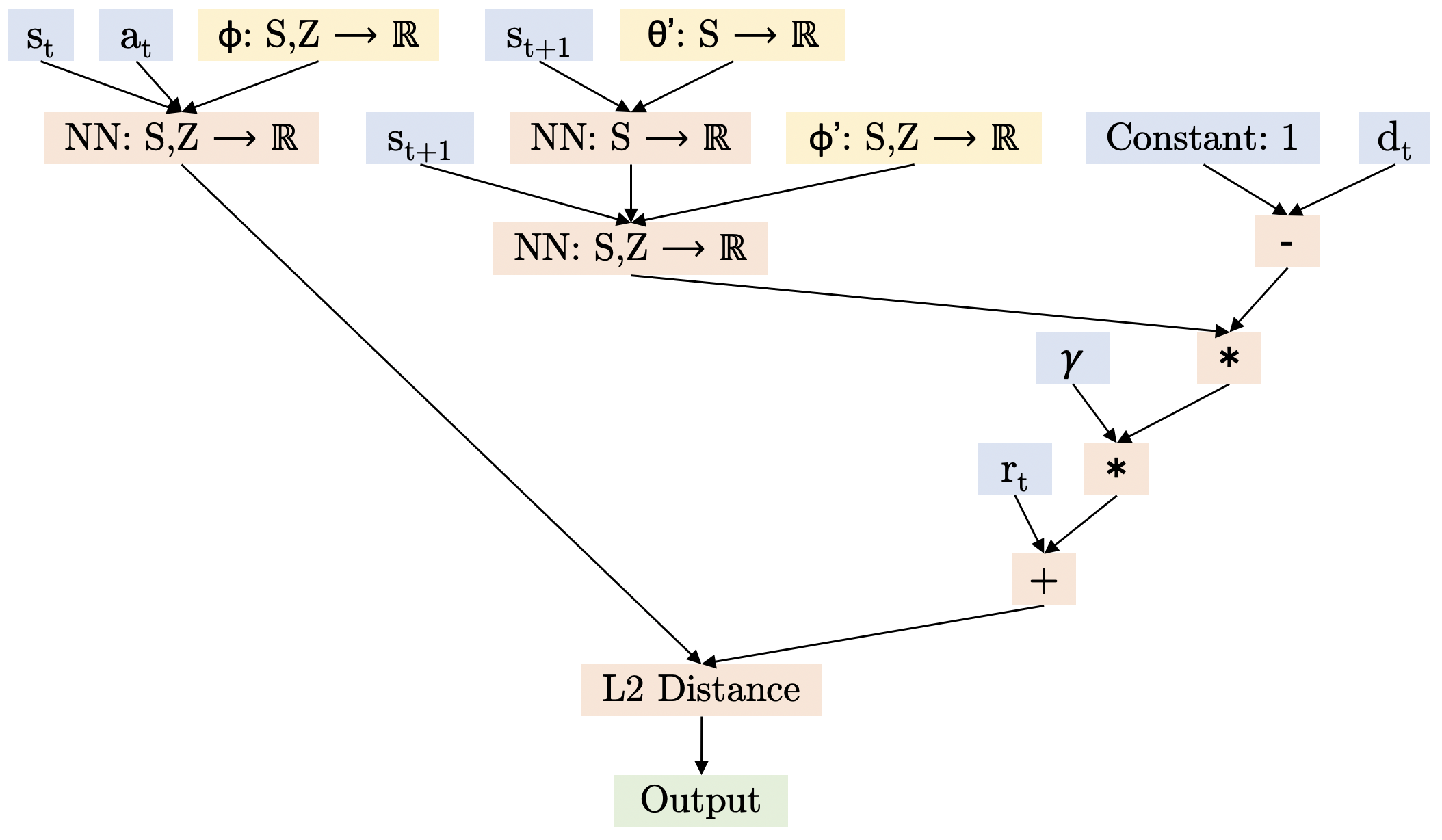}
\caption{Computational graph representing Deep Deterministic Policy Gradient (DDPG) Q-function $Q_{\phi}$ loss $L_{\phi} = \left(Q_{\phi}\left(s_t,a_t\right) - \left(r_t + \gamma\left(1-d_t\right)Q_{\phi'}\left(s_{t+1}, \mu_{\theta'}\left(s_{t+1}\right)\right)\right)\right)^2$.}
\label{fig:ddpg_phi}
\end{figure}

Next, we examine the DDPG algorithm, which in the original paper \cite{lillicrap2015continuous} was proposed to address continuous action spaces. As explained in the previous section, we adapt the framework to satisfy continuity requirements by introducing the parameter node $NN:\mathbb{S},\mathbb{Z}\longrightarrow\mathbb{R}$, which accounts for a Q-function taking a state and a continuous action as inputs. DDPG algorithm considers both a policy $\mu_{\theta}$ and a Q-function $Q_{\phi}$, as well as their target counterparts, $\mu_{\theta'}$ and $Q_{\phi'}$, respectively. Figure \ref{fig:ddpg_phi} shows the computational graph to update the Q-function parameters $\phi$, which makes use of the episode-termination signal $d_t$, also introduced in this work. 

\begin{figure}[!h]
\centering
\includegraphics[width=.32\linewidth]{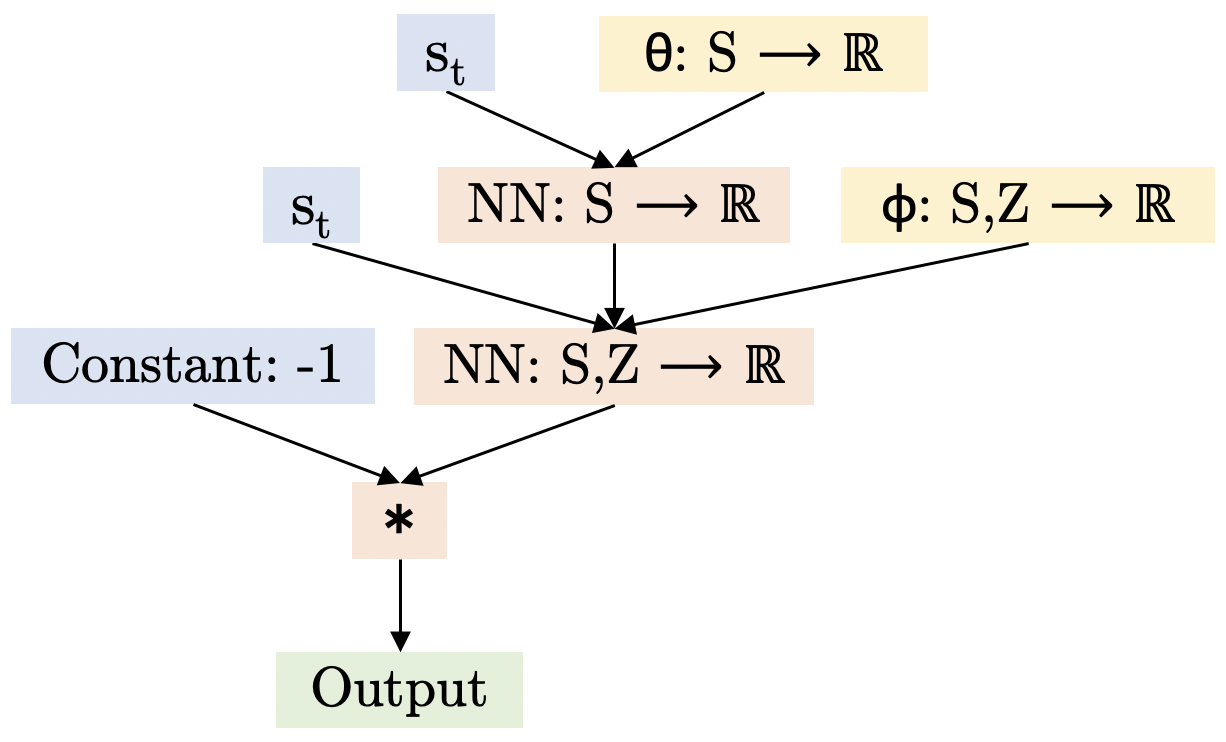}
\caption{Computational graph representing Deep Deterministic Policy Gradient (DDPG) policy $\mu_{\theta}$ loss $L_{\theta} = -Q_{\phi}\left(s_t, \mu_{\theta}\left(s_t\right)\right)$.}
\label{fig:ddpg_theta}
\end{figure}

Then, Figure \ref{fig:ddpg_theta} shows the computational graph to update policy parameters $\theta$.

\subsection{Twin Delayed DDPG (TD3)}

\begin{figure}[!h]
\centering
\includegraphics[width=.5\linewidth]{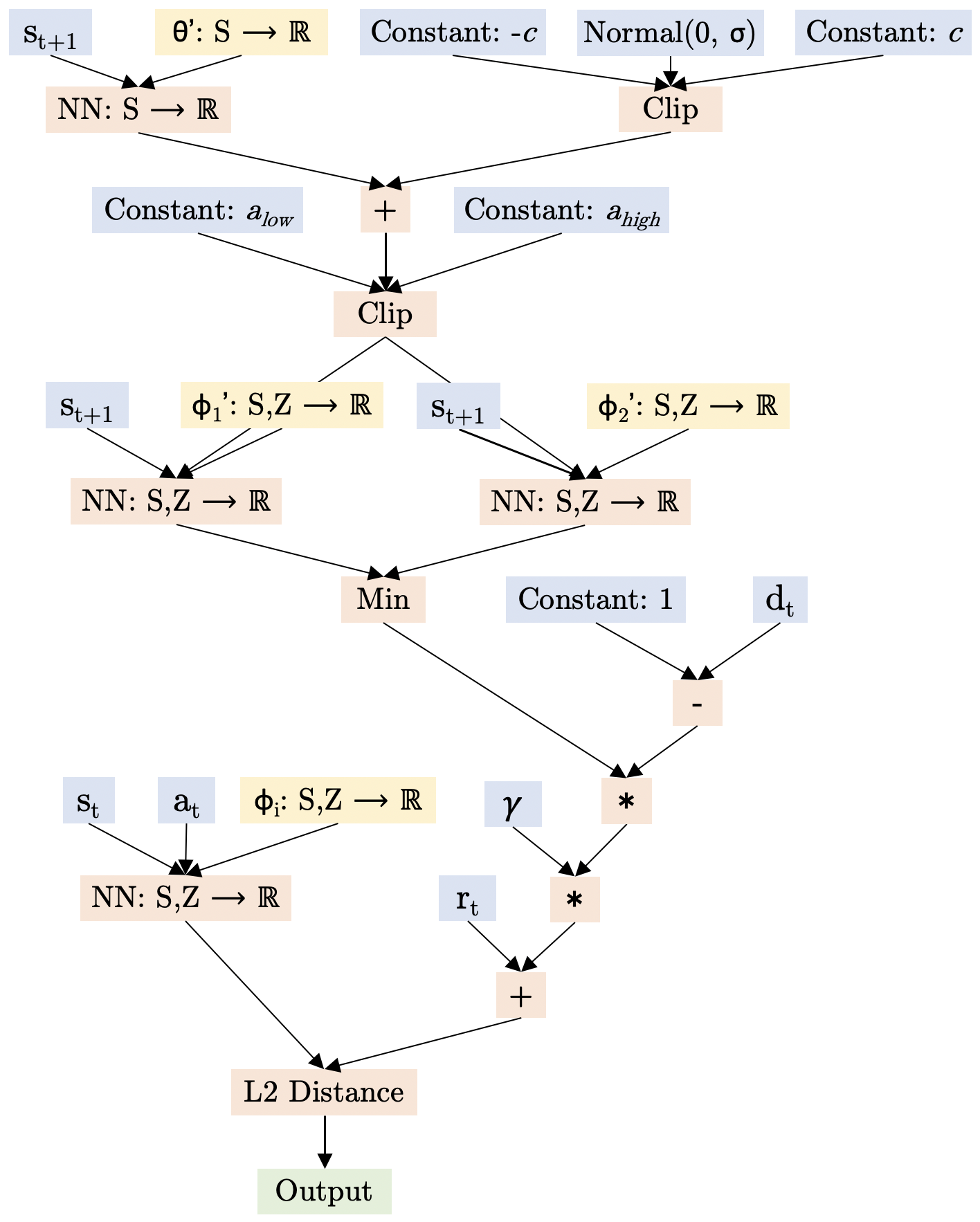}
\caption{Computational graph representing Twin Delayed DDPG (TD3) Q-function $Q_{\phi_i}$ loss, for $i=1,2$, given by $L_{\phi_i} = \left(Q_{\phi_i}\left(s_t, a_t\right) - \left(r_t + \gamma\left(1 - d_t\right)\underset{i=1,2}{\min}Q_{\phi'_i}\left(s_{t+1}, a\left(s_{t+1}\right)\right)\right)\right)^2$, where $a(s_{t+1}) = \text{clip}(\mu_{\theta'}(s_{t+1}) + \text{clip}(\epsilon, -c, c), a_{\text{low}}, a_{\text{high}})$, with $\epsilon\sim\mathcal{N}(0, \sigma)$.}
\label{fig:td3_phi}
\end{figure}

The TD3 loss functions can be constructed with all input, parameter, and operation nodes presented so far. However, new constant nodes are needed for these functions. Figure \ref{fig:td3_phi} shows the computational graph to update the Q-function parameters $\phi_i$. In the TD3 algorithm, clipped noise is added to the actions as a regularizer. In the graph, this is achieved by using input nodes with constant values $-c$ and $c$, as well as a constant $\epsilon$ sampled from the normal distribution $\mathcal{N}(0, \sigma)$. To make sure actions are kept within bounds, additional constant nodes with action bounds $a_{low}$ and $a_{high}$ are used.

\begin{figure}[!h]
\centering
\includegraphics[width=.32\linewidth]{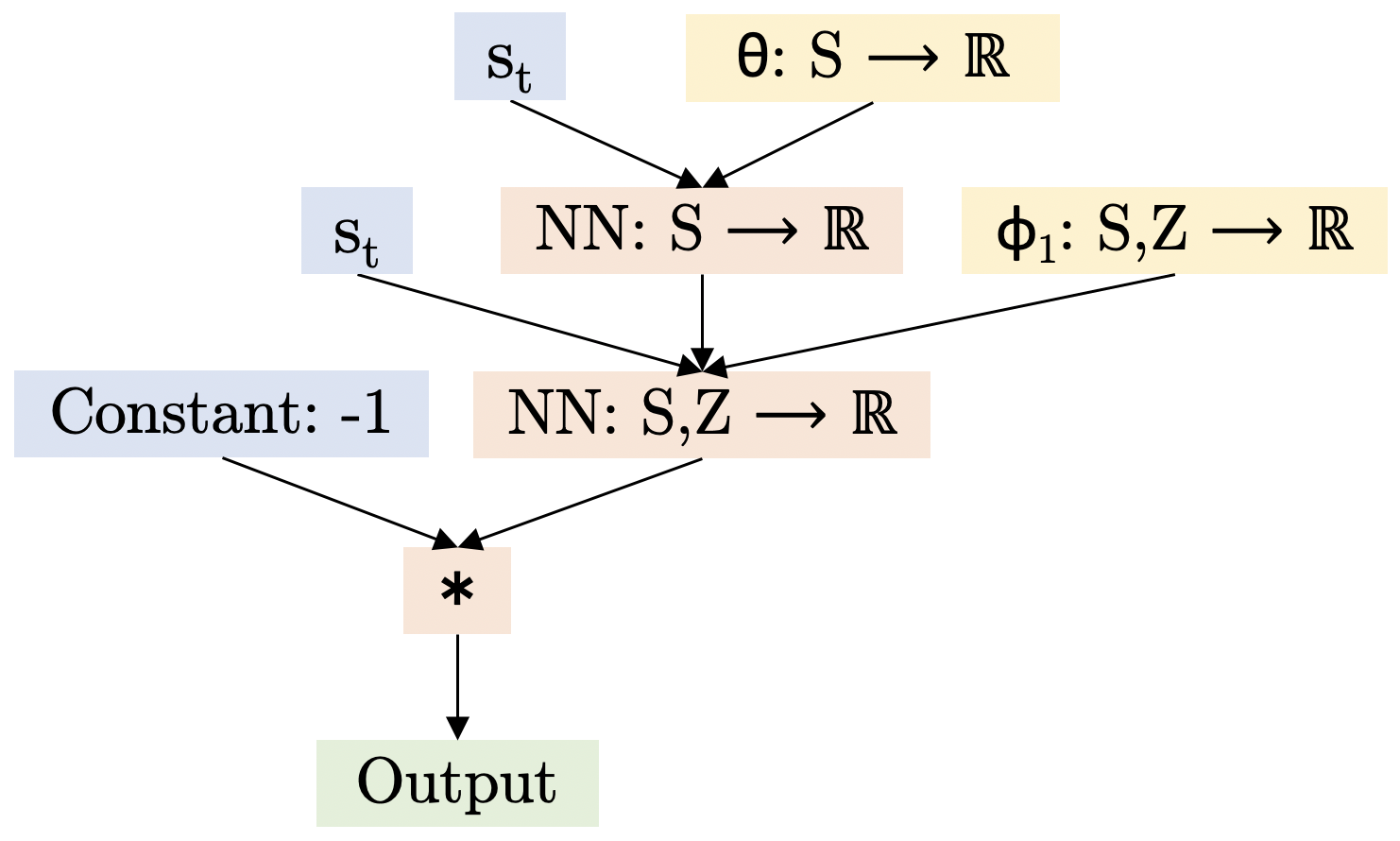}
\caption{Computational graph representing Twin Delayed DDPG (TD3) policy $\pi_{\theta}$ loss $L_{\theta} = - Q_{\phi_1}(s_t, \mu_{\theta}(s_t))$.}
\label{fig:td3_theta}
\end{figure}

Note this graph corresponds to generic Q-function parameters $\phi_i$; TD3 algorithm considers two different Q-functions, $Q_{\phi_1}$ and $Q_{\phi_2}$, the framework user can decide whether to use the same graph to update both or use two different DAGs. In contrast, only Q-function $Q_{\phi_1}$ is used to update policy parameters $\theta$, as shown in Figure \ref{fig:td3_theta}.

\subsection{Soft Actor Critic (SAC)}

\begin{figure}[!h]
\centering
\includegraphics[width=.5\linewidth]{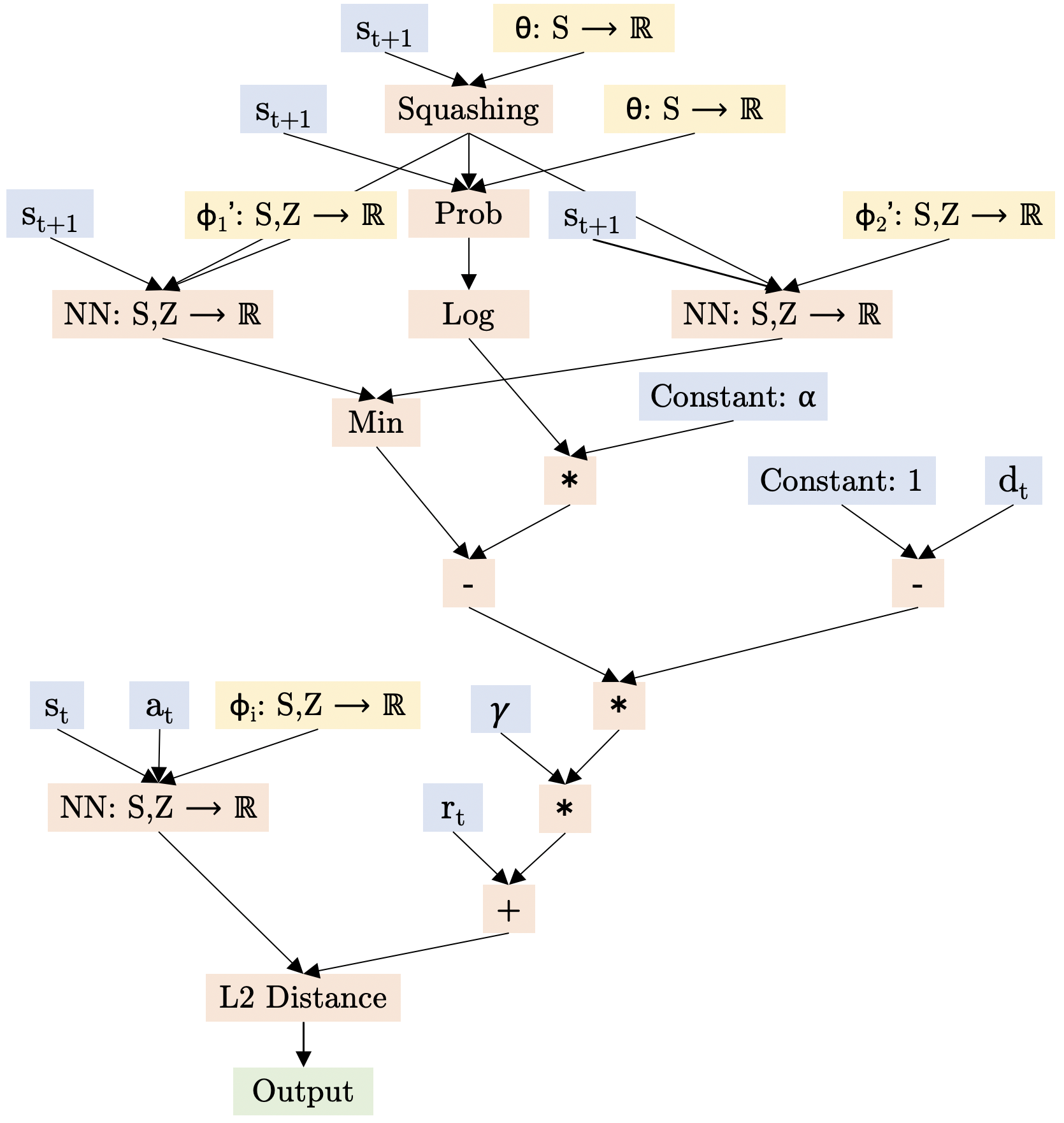}
\caption{Computational graph representing Soft Actor-Critic (SAC) Q-function $Q_{\phi}$ loss $L_{\phi} = \left(Q_{\phi_i}\left(s_t, a_t\right) - \left(r_t + \gamma\left(1 - d_t\right)\underset{i=1,2}{\min}Q_{\phi'_i}\left(s_{t+1}, \tilde{a}\right) - \alpha\log_{\pi_\theta}\left(\tilde{a}|s_{t+1}\right)\right)\right)^2$, where $\tilde{a} = \tanh(\mu_{\theta}(s_{t+1}) + \sigma_{\theta}(s_{t+1}) \odot \xi)$, with $\xi\sim\mathcal{N}(0, I)$.}
\label{fig:sac_phi}
\end{figure}

Finally, we address SAC loss functions. Following the implementation in \cite{SpinningUp2018}, we assume the use of environments with continuous action spaces. Figure \ref{fig:sac_phi} shows the computational graph corresponding to the generic Q-function $Q_{\phi_i}$ -- SAC uses two different Q-functions following the same rationale as TD3. In this case we use the two last operation nodes introduced in this work, \textit{Squashing} and \textit{Prob}. We follow the convention from the original paper of considering single-output nodes. In case the framework user wants to consider multiple-output nodes, the \textit{Squashing} and \textit{Prob} nodes could be integrated in a single node. To construct this graph, we also use an additional constant node with hyperparameter $\alpha$ \cite{haarnoja2018soft}.

\begin{figure}[!h]
\centering
\includegraphics[width=.4\linewidth]{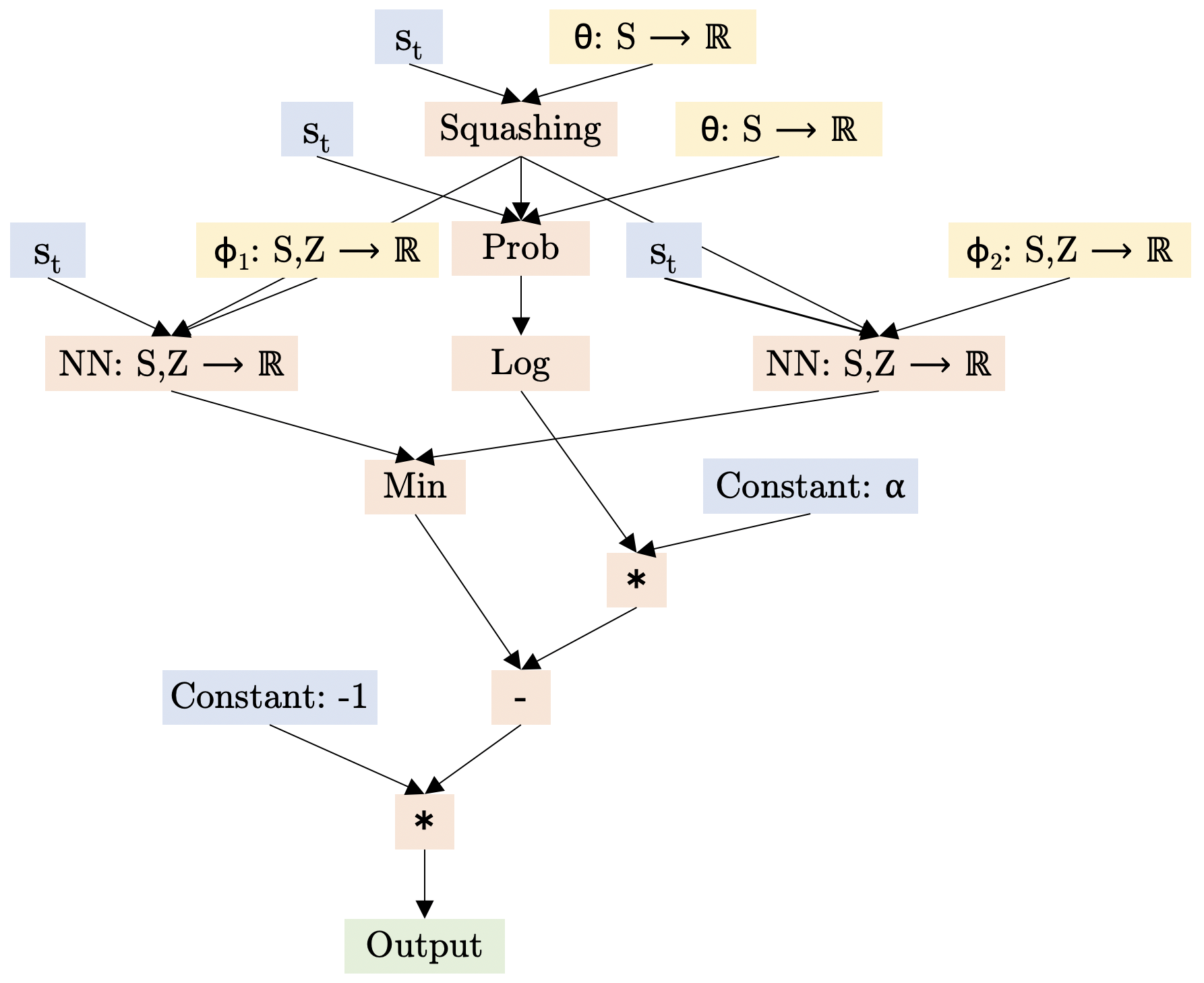}
\caption{Computational graph representing Soft Actor-Critic (SAC) policy $\pi_{\theta}$ loss $L_{\theta} = - \left(\underset{i=1,2}{\min}Q_{\phi_i}(s_t, \tilde{a})-\alpha\log_{\pi_\theta}(\tilde{a}|s_t)\right)$, where $\tilde{a} = \tanh(\mu_{\theta}(s_t) + \sigma_{\theta}(s_t) \odot \xi)$, with $\xi\sim\mathcal{N}(0, I)$.}
\label{fig:sac_theta}
\end{figure}

Figure \ref{fig:sac_theta} shows the computational graph to update policy parameters $\theta$, which makes use of the same new nodes.

\section{Conclusions}

In this work we have presented an extension of the search language introduced in \cite{coreyes2021evolving}, which serves as a set of building blocks to generate multiple RL algorithms in the form of Directed Acyclic Graphs. Specifically, we have focused on adapting the language to Policy Gradient algorithms and, to that end, proposed four additional operation nodes as well as five modifications to the input structure of the graphs. By means of these extensions, we have shown computational graphs for five different Policy Gradient algorithms: VPG, PPO, DDPG, TD3, and SAC.

Overall, representing RL algorithms as graphs offers high interpretability to the framework users, especially when RL algorithms are novel. As part of the future work, these graphs should be tested as warm starts to the original evolutionary framework and the resulting child graphs validated in different unseen environments. In addition, other RL algorithms not addressed in this work could be represented in the same format.

\section{Github Repository}

A Github repository with all of the figures shown in this work can be found at \href{https://github.com/jjgarau/DAGPolicyGradient}{https://github.com/jjgarau/DAGPolicyGradient}.



\bibliography{example_paper}

\begin{thebibliography}{10}

\bibitem{coreyes2021evolving}
John~D. Co-Reyes, Yingjie Miao, Daiyi Peng, Esteban Real, Sergey Levine,
  Quoc~V. Le, Honglak Lee, and Aleksandra Faust.
\newblock Evolving reinforcement learning algorithms, 2021.

\bibitem{1606.01540}
Greg Brockman, Vicki Cheung, Ludwig Pettersson, Jonas Schneider, John Schulman,
  Jie Tang, and Wojciech Zaremba.
\newblock Openai gym, 2016.

\bibitem{gym_minigrid}
Maxime Chevalier-Boisvert, Lucas Willems, and Suman Pal.
\newblock Minimalistic gridworld environment for openai gym.
\newblock \url{https://github.com/maximecb/gym-minigrid}, 2018.

\bibitem{mnih2015human}
Volodymyr Mnih, Koray Kavukcuoglu, David Silver, Andrei~A Rusu, Joel Veness,
  Marc~G Bellemare, Alex Graves, Martin Riedmiller, Andreas~K Fidjeland, Georg
  Ostrovski, et~al.
\newblock Human-level control through deep reinforcement learning.
\newblock {\em nature}, 518(7540):529--533, 2015.

\bibitem{vanhasselt2015deep}
Hado van Hasselt, Arthur Guez, and David Silver.
\newblock Deep reinforcement learning with double q-learning, 2015.

\bibitem{Bellemare_2013}
M.~G. Bellemare, Y.~Naddaf, J.~Veness, and M.~Bowling.
\newblock The arcade learning environment: An evaluation platform for general
  agents.
\newblock {\em Journal of Artificial Intelligence Research}, 47:253–279,
  2013.

\bibitem{schulman2017proximal}
John Schulman, Filip Wolski, Prafulla Dhariwal, Alec Radford, and Oleg Klimov.
\newblock Proximal policy optimization algorithms, 2017.

\bibitem{sutton2018reinforcement}
Richard~S Sutton and Andrew~G Barto.
\newblock {\em Reinforcement learning: An introduction}.
\newblock MIT press, 2018.

\bibitem{real2019regularized}
Esteban Real, Alok Aggarwal, Yanping Huang, and Quoc~V Le.
\newblock Regularized evolution for image classifier architecture search, 2019.

\bibitem{alet2020metalearning}
Ferran Alet, Martin~F. Schneider, Tomas Lozano-Perez, and Leslie~Pack
  Kaelbling.
\newblock Meta-learning curiosity algorithms, 2020.

\bibitem{houthooft2018evolved}
Rein Houthooft, Richard~Y. Chen, Phillip Isola, Bradly~C. Stadie, Filip Wolski,
  Jonathan Ho, and Pieter Abbeel.
\newblock Evolved policy gradients, 2018.

\bibitem{Todorov2012MuJoCoAP}
E.~Todorov, T.~Erez, and Y.~Tassa.
\newblock Mujoco: A physics engine for model-based control.
\newblock {\em 2012 IEEE/RSJ International Conference on Intelligent Robots and
  Systems}, pages 5026--5033, 2012.

\bibitem{bechtle2020metalearning}
Sarah Bechtle, Artem Molchanov, Yevgen Chebotar, Edward Grefenstette, Ludovic
  Righetti, Gaurav Sukhatme, and Franziska Meier.
\newblock Meta-learning via learned loss, 2020.

\bibitem{kirsch2020improving}
Louis Kirsch, Sjoerd van Steenkiste, and Jürgen Schmidhuber.
\newblock Improving generalization in meta reinforcement learning using learned
  objectives, 2020.

\bibitem{oh2020discovering}
Junhyuk Oh, Matteo Hessel, Wojciech~M. Czarnecki, Zhongwen Xu, Hado van
  Hasselt, Satinder Singh, and David Silver.
\newblock Discovering reinforcement learning algorithms, 2020.

\bibitem{lillicrap2015continuous}
Timothy~P Lillicrap, Jonathan~J Hunt, Alexander Pritzel, Nicolas Heess, Tom
  Erez, Yuval Tassa, David Silver, and Daan Wierstra.
\newblock Continuous control with deep reinforcement learning.
\newblock {\em arXiv preprint arXiv:1509.02971}, 2015.

\bibitem{haarnoja2018soft}
Tuomas Haarnoja, Aurick Zhou, Pieter Abbeel, and Sergey Levine.
\newblock Soft actor-critic: Off-policy maximum entropy deep reinforcement
  learning with a stochastic actor, 2018.

\bibitem{fujimoto2018addressing}
Scott Fujimoto, Herke van Hoof, and David Meger.
\newblock Addressing function approximation error in actor-critic methods,
  2018.

\bibitem{SpinningUp2018}
Joshua Achiam.
\newblock {Spinning Up in Deep Reinforcement Learning}.
\newblock 2018.

\end{thebibliography}
\bibliographystyle{unsrt}

\end{document}